\documentclass[runningheads]{llncs}
\usepackage[T1]{fontenc}
\usepackage{graphicx}
\usepackage{multirow} 
\usepackage{makecell}
\usepackage{booktabs}
\usepackage{amsfonts}
\usepackage{amsmath}
\usepackage{ulem}
\usepackage{hyperref}
\usepackage{float}
\usepackage{marvosym}

\hypersetup{hidelinks}

\newfloat{figtab}{htb}{fgtb}
\makeatletter
  \newcommand\figcaption{\def\@captype{figure}\caption}
  \newcommand\tabcaption{\def\@captype{table}\caption}
\makeatother

\begin{document}
\title{HiMatch-AD: DINOv3-driven Hierarchical Matching for Training-free Medical Anomaly Detection}

\titlerunning{HiMatch-AD}

\author{Jiayu Huo\inst{1}\textsuperscript{(\Letter)} \and
Jingyuan Hong\inst{2} \and
Meng Zhou\inst{1} \and
Liyun Chen\inst{3} \and
Le Zhang\inst{4}
}

\authorrunning{J. Huo et al.}

\institute{Imperial College London, London, UK \\
\email{j.huo@ic.ac.uk}
\and
King’s College London, London, UK \\
\and
SonoScape Medical Corp. \\
\and
University of Birmingham, Birmingham, UK
}

\maketitle              
\begin{abstract}
Anomaly detection is essential for medical image analysis, where pathological regions often appear as rare deviations from normal anatomical structures. While training-based methods have achieved promising performance, they require task-specific optimization and extensive normal data, which limits scalability across modalities and institutions. Training-free approaches offer greater flexibility by leveraging pretrained visual representations, yet existing methods typically rely on simple nearest-neighbor retrieval and naive aggregation strategies, which may fail to capture hierarchical semantics and ignore the reliability of multiple anomaly responses. In this work, we propose HiMatch-AD, a DINOv3-driven hierarchical matching framework for training-free medical anomaly detection. Our method first retrieves semantically relevant normal references via dual-branch matching that jointly considers global CLS-token similarity and patch-level representations. Hierarchical anomaly maps are then generated across multiple transformer stages by comparing clustered normal features with query representations. To robustly aggregate anomaly responses, we introduce a unified uncertainty-based fusion mechanism that adaptively weights maps according to their reliability. The entire framework operates without any task-specific training. Extensive experiments on the BMAD benchmark, including brain MRI, liver CT, and retinal OCT datasets, demonstrate that HiMatch-AD consistently outperforms both training-based and DINO-based state-of-the-art methods, which highlights the effectiveness of multi-level matching and uncertainty-aware fusion for scalable medical anomaly detection.

\end{abstract}

\section{Introduction}

Anomaly detection (AD) plays a critical role in medical image analysis, as many clinically significant findings correspond to rare or abnormal patterns that deviate from normal anatomical structures~\cite{ammar2025foundation,cao2024survey}. Reliable anomaly detection tools can assist radiologists in identifying tumors, lesions, or pathological regions, particularly in scenarios where annotated abnormal data are scarce~\cite{cai2025medianomaly,fernando2021deep}. Compared to fully supervised segmentation, anomaly detection offers a more scalable solution, since collecting pixel-level annotations for diverse diseases is often costly and time-consuming. Therefore, developing robust and generalizable anomaly detection frameworks is of great importance for clinical deployment.

Existing anomaly detection approaches can be broadly categorized into training and training-free paradigms. Training-based methods, including student-teacher and reconstruction-based frameworks, have shown promising results on both natural and medical images~\cite{salehi2021multiresolution,deng2022anomaly,wang2021student,zavrtanik2021draem,chen2022utrad,cai2026raid,schlegl2019f}. However, they require task-specific training and often rely on carefully curated normal datasets, which limits their scalability across imaging modalities and institutions. In contrast, training-free methods~\cite{huo2026dino,damm2025anomalydino,schulthess2025anomaly,gu2025univad,roth2022towards}, such as memory-based matching approaches~\cite{damm2025anomalydino,schulthess2025anomaly,roth2022towards}, avoid optimization on target datasets and instead leverage pretrained representations for anomaly scoring. While more flexible, these methods typically depend on simple nearest-neighbor retrieval or naive aggregation strategies, which may fail to capture hierarchical semantic structures and often overlook the reliability of multiple anomaly responses.

To address these limitations, we propose \textbf{HiMatch-AD}, a DINOv3~\cite{simeoni2025dinov3}-driven multi-level matching framework for training-free medical anomaly detection. Our method first retrieves semantically relevant normal references via dual-branch matching that jointly considers global CLS-token similarity and spatially pooled patch-token similarity. Given the retrieved support set, we perform hierarchical anomaly map generation across multiple transformer stages, where normal feature distributions are summarized via clustering and compared against query features. To robustly aggregate anomaly responses, we introduce a unified uncertainty-based fusion mechanism that adaptively weights maps according to their reliability. The entire framework operates without task-specific training, leveraging strong pretrained visual representations to enable scalable anomaly detection.

In summary, our contributions are three-fold: (1) We design a dual-branch support retrieval strategy and hierarchical anomaly map generation scheme to capture both global and local feature discrepancies. (2) We introduce an uncertainty-aware fusion mechanism that improves robustness and localization accuracy. (3) Extensive experiments on the BMAD benchmark demonstrate that our proposed HiMatch-AD consistently outperforms state-of-the-art (SOTA) training-based and training-free methods.

\section{Methodology}

\begin{figure}[!t]
\centering
\includegraphics[width=0.98\textwidth]{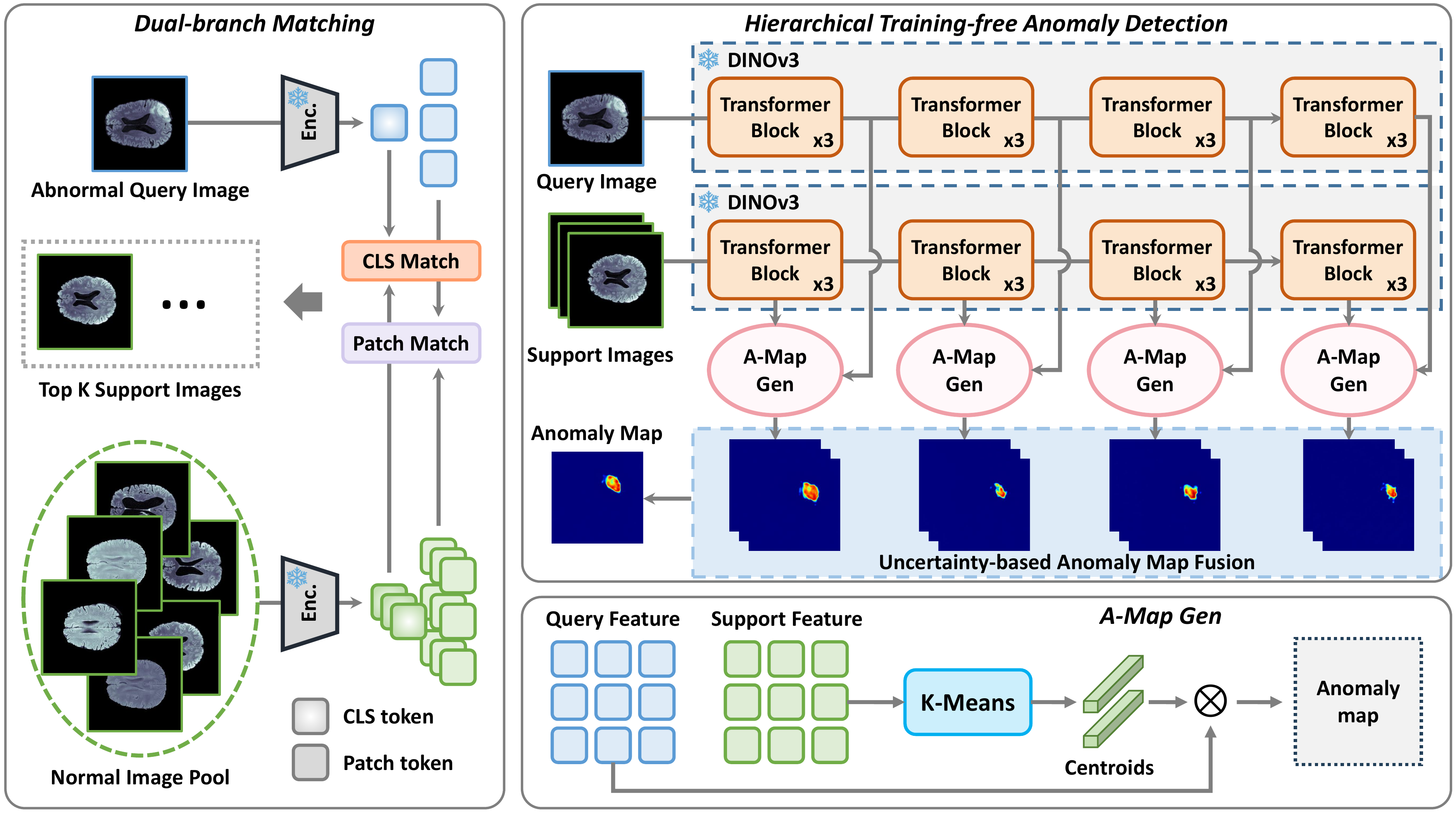}
\caption{Overview of the proposed HiMatch-AD framework. Our HiMatch-AD framework contains two steps: $K$ support images are retrieved through dual-branch matching (left). Then the query and support images are sent to the hierarchical training-free anomaly detection pipeline for anomaly map generation (upper right). The details of A-Map Gen module are shown on the bottom right. Here ``Enc.'' denotes the encoder, where DINOv3 is applied.}
\label{fig:main_framework}
\end{figure}

\subsection{Dual-branch Matching for Support Image Retrieval}
\label{section_dual_branch_matching}
Given an abnormal query image $x^q$ and a normal image pool $\mathcal{X}^n = \{x_i^n\}_{i=1}^{N}$, 
we first encode all images using a pretrained DINOv3 vision transformer with a patch size of 16. 
For each image, we extract the output tokens from the last transformer block, 
including the global \texttt{[CLS]} token and the patch tokens.

Let $\mathbf{f}_{\mathrm{cls}} \in \mathbb{R}^{d}$ denote the CLS token representation, 
and $\mathbf{F}_{\mathrm{patch}} \in \mathbb{R}^{M \times d}$ denote the patch token representations, 
where $M$ is the number of spatial patches and $d$ is the embedding dimension. 
To obtain a compact global descriptor for the patch branch, 
we perform spatial average pooling over all patch tokens:

\begin{equation}
\mathbf{f}_{\mathrm{patch}} = \frac{1}{M} \sum_{m=1}^{M} \mathbf{F}_{\mathrm{patch}}^{(m)} .
\end{equation}

For the query image $x^q$ and each normal image $x_i^n$, 
we compute cosine similarities separately for the CLS branch and the patch branch:

\begin{equation}
s_{\mathrm{cls}}(x^q, x_i^n) = 
\frac{\mathbf{f}_{\mathrm{cls}}^q \cdot \mathbf{f}_{\mathrm{cls}}^i}
{\|\mathbf{f}_{\mathrm{cls}}^q\|_2 \|\mathbf{f}_{\mathrm{cls}}^i\|_2},
s_{\mathrm{patch}}(x^q, x_i^n) = 
\frac{\mathbf{f}_{\mathrm{patch}}^q \cdot \mathbf{f}_{\mathrm{patch}}^i}
{\|\mathbf{f}_{\mathrm{patch}}^q\|_2 \|\mathbf{f}_{\mathrm{patch}}^i\|_2}.
\end{equation}

The final similarity score $s(x^q, x_i^n)$ is obtained by aggregating the two branches. We then select the top-$K$ normal images with the highest similarity scores 
to construct the support set $\mathcal{S}(x^q)$:

\begin{equation}
\mathcal{S}(x^q) = \operatorname{TopK}_{x_i^n \in \mathcal{X}^n} \; s(x^q, x_i^n).
\end{equation}

The retrieved support images serve as normal references for subsequent anomaly map generation.

\subsection{Computation of Anomaly Maps}
\label{section_am_computation}
Given a query image $x^q$ and its retrieved support set 
$\mathcal{S}(x^q)=\{x_k^s\}_{k=1}^{K}$, 
we feed both the query and support images into the same pretrained DINOv3 encoder. 
The backbone consists of 12 transformer blocks, and we group every three consecutive blocks 
into one stage, resulting in four hierarchical stages. For each stage $t \in \{1,2,3,4\}$, we extract the corresponding feature maps: $\mathbf{Q}^{(t)} \in \mathbb{R}^{M \times d}$, and 
$\mathbf{S}_k^{(t)} \in \mathbb{R}^{M \times d}$. Here $\mathbf{Q}^{(t)}$ and $\mathbf{S}_k^{(t)}$ represent the patch-level features of the query image and the $k$-th support image at stage $t$, respectively.

For each stage $t$ and each support image $x_k^s$, we generate an anomaly map via the A-Map Gen module. First, we apply K-means clustering to the support features $\mathbf{S}_k^{(t)}$ to obtain $C$ representative centroids: $\{\mathbf{c}_{k,j}^{(t)}\}_{j=1}^{C}, \mathbf{c}_{k,j}^{(t)} \in \mathbb{R}^{1 \times d}.$ These centroids summarize the normal feature distribution of the support image at the current stage. Next, for each centroid $\mathbf{c}_{k,j}^{(t)}$, we compute the cosine similarity between the centroid and each query feature $\mathbf{Q}^{(t)}$:

\begin{equation}
a_{k,j}^{(t)} = 
\frac{\mathbf{Q}^{(t)} \cdot \mathbf{c}_{k,j}^{(t)}}
{\|\mathbf{Q}^{(t)}\|_2 \|\mathbf{c}_{k,j}^{(t)}\|_2}.
\end{equation}

This produces $C$ initial anomaly maps for each support image at stage $t$. To convert similarity $a_{k,j}^{(t)}$ into anomaly responses $\tilde{a}_{k,j}^{(t)}$, we calculate $\tilde{a}_{k,j}^{(t)} = 1 - (a_{k,j}^{(t)} + 1) / 2$, such that larger values indicate higher abnormality. To aggregate the responses from multiple centroids, we adopt a softmax fusion strategy over the centroid dimension:

\begin{equation}
A_k^{(t)} = 
\sum_{j=1}^{C} 
\frac{\exp(\tilde{a}_{k,j}^{(t)})}
{\sum_{j'=1}^{C} \exp(\tilde{a}_{k,j'}^{(t)})} 
\, \tilde{a}_{k,j}^{(t)}.
\end{equation}

The softmax operation assigns larger weights to higher anomaly responses, thereby emphasizing query regions that are dissimilar to the normal support distribution. The resulting $A_k^{(t)} \in \mathbb{R}^{M}$ is the anomaly map corresponding to the $k$-th support image at stage $t$. This procedure is performed hierarchically across all stages to capture anomalies at different semantic levels.

\subsection{Uncertainty-based Anomaly Map Fusion}
\label{section_uncertainty_fusion}
We adopt a unified uncertainty-aware fusion strategy to aggregate multiple anomaly maps. Without loss of generality, we first describe the instance-wise fusion, where anomaly maps originate from different support images. The same formulation is later applied to stage-wise fusion.

Given a set of anomaly maps $\{A_i\}_{i=1}^{K}$ obtained from $K$ support images, where $A_i \in \mathbb{R}^{M}$. we first compute the mean anomaly map:

\begin{equation}
\bar{A} = 
\frac{1}{K} \sum_{i=1}^{K} A_i.
\end{equation}

Based on the mean response, we quantify the uncertainty of each anomaly map as its deviation from the mean by calculating $U_i = \left(A_i - \bar{A}\right)^2$. Higher deviation indicates lower reliability. We therefore compute fusion weights using a softmin operator over the uncertainty dimension:

\begin{equation}
w_i = 
\frac{\exp\left(-U_i\right)}
{\sum_{j=1}^{N} \exp\left(-U_j\right)}.
\end{equation}

The fused anomaly map is obtained as:

\begin{equation}
A^{\mathrm{fused}} = 
\sum_{i=1}^{N} w_i \, A_i.
\end{equation}

The stage-wise fusion follows the same formulation, where $\{A_i\}_{i=1}^{N}$ corresponds to anomaly maps generated from $N$ hierarchical stages instead of different support instances. The unified uncertainty-based formulation enables adaptive weighting of anomaly responses regardless of their origin.

\section{Experiments}
\subsection{Dataset}
We evaluate HiMatch-AD on three medical anomaly detection benchmarks from the BMAD~\cite{bao2024bmad} benchmark, including brain MRI, liver CT, and retinal OCT datasets. The brain MRI dataset is derived from the BraTS 2021~\cite{baid2021rsna} dataset. It comprises 7,500 normal images for training, while the test set includes 640 normal images and 3,075 abnormal images annotated with pixel-level anomaly masks. The liver CT dataset is constructed based on the BTCV~\cite{landman2015miccai} and LiTs~\cite{bilic2023liver} datasets. The training split consists of 1,542 normal images. For evaluation, the test set contains 833 normal images and 660 abnormal images with corresponding segmentation masks. The retinal OCT dataset is obtained from the RESC~\cite{hu2019automated} dataset. Its training set includes 4,297 normal images, whereas the test set is composed of 1,041 normal images and 764 abnormal images with ground-truth masks. For all datasets, only normal training images are used for support set retrieval in the proposed training-free framework.

\subsection{Implementation Details}
We adopt DINOv3 ViT-B/16~\cite{simeoni2025dinov3,dosovitskiy2020image} as the feature extraction backbone. All backbone parameters remain frozen during inference, and no task-specific training is performed in our framework. We evaluate the proposed method using three standard metrics: the Area Under the Receiver Operating Characteristic Curve for image-level anomaly detection (I-AUC), pixel-level anomaly localization (P-AUC), and pixel-level Per-Region Overlap (P-PRO). Since abnormal regions typically occupy only a small fraction of each image, with the majority remaining normal, we follow a maximum-response strategy to compute image-level anomaly scores. Specifically, the maximum value of the pixel-level anomaly map is used as the image-level anomaly score for I-AUC evaluation. We set the number of centroids $C$ to 2 according to~\cite{huo2026dino}.

\section{Results}
\subsection{Comparisons to SOTA Anomaly Detection Models}
We compare HiMatch-AD with a broad range of SOTA anomaly detection methods. The training-based approaches include DRAEM~\cite{zavrtanik2021draem}, UTRAD~\cite{chen2022utrad}, SimpleNet~\cite{liu2023simplenet}, MKD~\cite{salehi2021multiresolution}, RD4AD~\cite{deng2022anomaly}, STFPM~\cite{wang2021student}, and PaDiM~\cite{defard2021padim}. We also compare against training-free or memory-based methods, including PatchCore~\cite{roth2022towards}, as well as recent DINO-based approaches, AnomalyDINO~\cite{damm2025anomalydino} and DINO-DPMM~\cite{schulthess2025anomaly}.

Table~\ref{tab:main_results} presents the quantitative comparison on the BMAD benchmark across brain MRI, liver CT, and retinal OCT datasets. Overall, HiMatch-AD achieves the best performance across all datasets and evaluation metrics. On the brain MRI dataset, our method attains the highest I-AUC (92.91\%), P-AUC (98.87\%), and P-PRO (87.73\%), surpassing both training-based and DINO-based competitors. Compared with the strongest baseline AnomalyDINO, our approach improves I-AUC by 1.10\% and P-AUC by 1.64\%. On the liver CT dataset, HiMatch-AD achieves 77.84\% I-AUC and 98.92\% P-AUC, outperforming all competing methods. Notably, compared with the second-best DINO-based method, our approach improves I-AUC by 2.41\% over AnomalyDINO and achieves the highest P-PRO (92.75\%). For the retinal OCT dataset, HiMatch-AD consistently delivers the best results across all metrics, achieving 92.61\% I-AUC, 97.01\% P-AUC, and 86.23\% P-PRO. It outperforms PatchCore and AnomalyDINO, demonstrating the effectiveness of hierarchical matching and uncertainty-aware fusion. Importantly, HiMatch-AD achieves these improvements without any task-specific training, highlighting the advantage of our training-free multi-level matching framework. The superior performance across all datasets demonstrates that leveraging strong pretrained representations with principled matching and fusion strategies can bridge the gap between training-based and training-free anomaly detection.

\begin{table*}[t]
\centering
\caption{Comparisons to SOTA models on BMAD benchmark datasets. The best results are in bold and the second best are underlined.}
\label{tab:main_results}
\fontsize{9}{11}\selectfont
\begin{tabular}{l|ccc|ccc|ccc}
\hline
\multirow{2}{*}{Model} 
& \multicolumn{3}{c|}{Brain MRI} 
& \multicolumn{3}{c|}{Liver CT} 
& \multicolumn{3}{c}{Retinal OCT} \\

& I-AUC & P-AUC & P-PRO 
& I-AUC & P-AUC & P-PRO 
& I-AUC & P-AUC & P-PRO \\
\hline

DRAEM      & 62.35 & 82.29 & 63.76 & 69.95 & 87.45 & 79.29 & 83.22 & 86.79 & 63.55 \\
UTRAD      & 82.92 & 92.61 & 72.29 & 55.81 & 87.88 & 71.12 & 89.39 & 94.54 & 77.49 \\
SimpleNet  & 82.52 & 94.76 & 78.38 & 72.28 & \underline{97.51} & 91.07 & 76.15 & 77.14 & 49.07 \\
MKD        & 81.47 & 89.44 & 67.59 & 60.72 & 96.06 & \underline{91.08} & 89.00 & 86.74 & 66.17 \\
RD4AD      & 89.45 & 96.45 & 85.86 & 60.38 & 96.01 & 90.29 & 87.77 & 96.18 & 85.62 \\
STFPM      & 83.04 & 95.62 & 83.02 & 61.75 & 91.18 & 90.62 & 84.82 & 94.68 & 81.27 \\
PaDiM      & 79.02 & 94.37 & 76.41 & 50.78 & 90.94 & 76.79 & 75.87 & 91.44 & 71.68 \\
PatchCore  & 91.65 & 96.97 & 85.68 & 60.28 & 96.43 & 87.75 & \underline{91.55} & \underline{96.48} & \underline{85.84} \\
\hline
AnomalyDINO & \underline{91.81} & \underline{97.23} & \underline{86.21} & \underline{75.43} & 97.12 & 91.01 & 90.19 & 94.04 & 84.24 \\
DINO-DPMM   & 90.73 & 96.02 & 85.74 & 73.47 & 93.47 & 90.74 & 89.45 & 90.20 & 83.47 \\
\hline
Ours & \textbf{92.91} & \textbf{98.87} & \textbf{87.73} 
     & \textbf{77.84} & \textbf{98.92} & \textbf{92.75} 
     & \textbf{92.61} & \textbf{97.01} & \textbf{86.23} \\
\hline
\end{tabular}
\end{table*}

\subsection{Ablation Studies}

\begin{figure}[!t]
\centering
\includegraphics[width=0.98\textwidth]{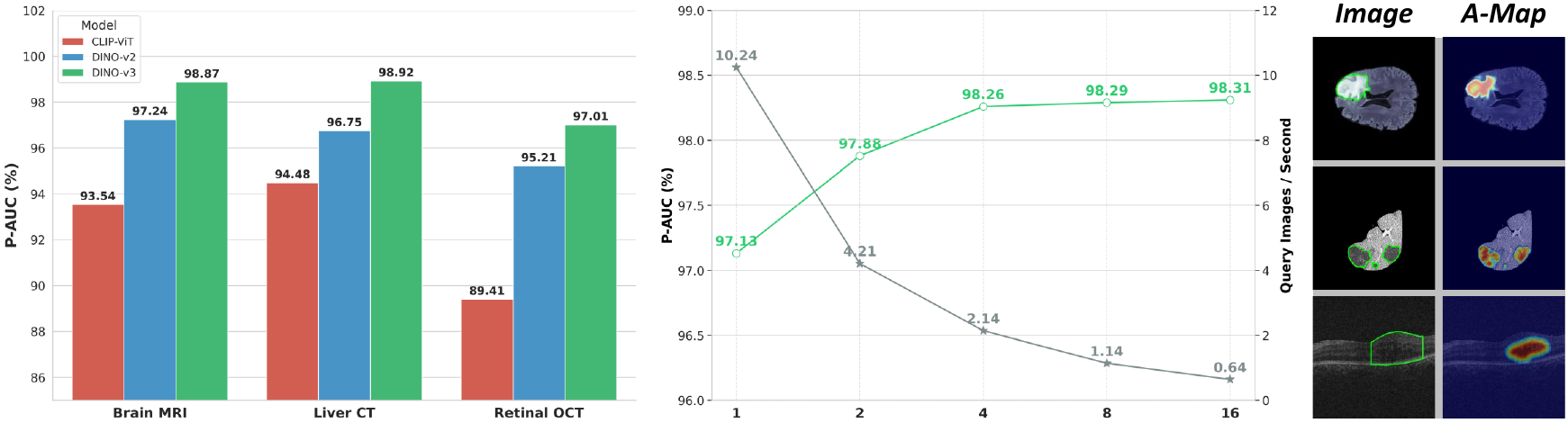}
\caption{Quantitative comparison of different ViT backbones for feature extraction (left). Performance and inference speed as the number of support images $K$ increases from 1 to 16 (middle). Visualization of anomaly maps generated by our approach, with green contours indicating anomalous regions (right).} 
\label{fig:ablation_study_bar_line_plot}
\end{figure}

We conduct comprehensive ablation studies to analyze the impact of backbone selection, the number of support images $K$, the matching and fusion strategies.

\subsubsection{Backbone and Support Size Analysis.}
We first investigate the effect of different pretrained ViT models for feature extraction. As shown in Fig.~\ref{fig:ablation_study_bar_line_plot} (left), DINOv3 consistently outperforms CLIP-ViT~\cite{radford2021learning} and DINOv2~\cite{oquab2023dinov2} across all three datasets. For example, on brain MRI, DINOv3 achieves 98.87\% P-AUC, compared to 97.24\% (DINOv2) and 93.54\% (CLIP-ViT). Similar improvements are observed on liver CT (98.92\%) and retinal OCT (97.01\%), demonstrating the advantage of stronger self-supervised representations for feature matching. We further analyze the influence of the support set size $K$. As illustrated in Fig.~\ref{fig:ablation_study_bar_line_plot} (middle), increasing $K$ from 1 to 4 markedly improves performance (from 97.13\% to 98.26\% P-AUC). However, further increasing $K$ yields only marginal gains (98.29\% at $K=8$ and 98.31\% at $K=16$), while inference speed drops notably (from 10.24 to 0.64 query images per second). Considering this trade-off, we set $K=4$ for all experiments. We also show anomaly maps generated by our approach in Fig.~\ref{fig:ablation_study_bar_line_plot} (right), illustrating strong alignment between high-response regions and ground-truth abnormalities (green contour) across different imaging modalities.

\subsubsection{Effect of Matching Strategy.}
We next evaluate different support retrieval strategies, as shown in Table~\ref{tab:ablation_study}. If neither CLS matching nor Patch matching is selected, the support set is randomly sampled from the normal pool. Random selection leads to substantially degraded performance (85.54\% / 82.20\% / 83.42\% on brain MRI / liver CT / retinal OCT), highlighting the importance of similarity-based retrieval. Using only the CLS branch improves results markedly, while using only the Patch branch achieves further gains. Combining both branches consistently delivers the best performance (98.87\%, 98.92\%, and 97.01\% respectively), demonstrating that global semantic alignment and local patch-level similarity provide complementary information.

\subsubsection{Effect of Fusion Strategy.}
We also analyze the impact of different fusion mechanisms. If neither Stage nor Instance fusion is enabled, anomaly maps are combined via simple averaging. Average fusion yields inferior results (96.42\% / 97.05\% / 94.84\%), indicating that naive aggregation fails to account for reliability differences. Applying either stage-wise or instance-wise uncertainty fusion improves performance, while jointly enabling both achieves the best results (98.87\%, 98.92\%, and 97.01\%). This confirms that uncertainty-aware weighting effectively suppresses inconsistent anomaly responses and enhances robust localization.

Overall, these ablations validate the necessity of dual-branch matching and unified uncertainty-based fusion in the proposed HiMatch-AD framework.

\begin{table*}[!t]
\centering
\caption{Ablation studies of the matching method and the fusion method of our proposed method HiMatch-AD. We keep the number of the support set $K$ to 4 across all ablation experiments.}
\label{tab:ablation_study}
\fontsize{9}{11}\selectfont

\setlength{\tabcolsep}{6pt}

\begin{tabular}{
>{\centering\arraybackslash}m{1.1cm}
>{\centering\arraybackslash}m{1.1cm}
|c|c|c}
\hline
\multicolumn{2}{c|}{Matching Method} 
& \multirow{2}{*}{Brain MRI} 
& \multirow{2}{*}{Liver CT} 
& \multirow{2}{*}{Retinal OCT} \\

CLS & Patch & & & \\
\hline

             &              & 85.54 & 82.20 & 83.42  \\
$\checkmark$ &              & 96.23 & 96.57 & 95.72  \\
             & $\checkmark$ & 97.46 & 97.41 & 96.75  \\
$\checkmark$ & $\checkmark$ & \textbf{98.87} & \textbf{98.92} & \textbf{97.01}  \\
\hline
\hline

\multicolumn{2}{c|}{Fusion Method} 
& \multirow{2}{*}{Brain MRI} 
& \multirow{2}{*}{Liver CT} 
& \multirow{2}{*}{Retinal OCT} \\

Stage & Instance & & & \\
\hline

             &              & 96.42 & 97.05 & 94.84  \\
$\checkmark$ &              & 97.54 & 97.84 & 95.41  \\
             & $\checkmark$ & 98.11 & 98.15 & 96.62  \\
$\checkmark$ & $\checkmark$ & \textbf{98.87} & \textbf{98.92} & \textbf{97.01}  \\
\hline
\end{tabular}

\end{table*}

\section{Conclusion}
In this work, we propose HiMatch-AD, a training-free anomaly detection framework based on DINOv3-driven multi-level matching. By integrating dual-branch support retrieval with hierarchical anomaly map generation, our method effectively captures both global semantic consistency and local feature deviations. Furthermore, we introduce a unified uncertainty-based fusion strategy that adaptively aggregates anomaly responses across support instances and hierarchical stages. Extensive experiments on the BMAD benchmark, including brain MRI, liver CT, and retinal OCT datasets, demonstrate that HiMatch-AD consistently outperforms both training-based and DINO-based SOTA methods. The ablation studies further validate the importance of dual-branch matching and uncertainty-aware fusion in improving robustness and localization accuracy. Without requiring any task-specific training, HiMatch-AD provides a scalable and efficient paradigm for medical anomaly detection. In future work, we plan to explore more advanced distribution modeling techniques and extend the proposed framework to broader multi-modal medical imaging scenarios.

\bibliographystyle{splncs04}
\bibliography{books}

\end{document}